%% file: wepa_bm.tex
\documentclass[runningheads]{llncs}
\usepackage[T1]{fontenc}
\usepackage{graphicx}
\usepackage{amsmath}
\usepackage{amssymb}
\usepackage{tabularx}
\usepackage{nicematrix}
\usepackage{booktabs}
\usepackage{subfig}
\usepackage{makecell}
\usepackage{siunitx}
\usepackage{multirow}
\usepackage{graphicx}
\newcolumntype{C}{>{\centering\arraybackslash}X}

\usepackage[export]{adjustbox} % for valign option

\usepackage{enumitem}

\makeatletter
\renewcommand\paragraph{\@startsection{paragraph}{4}{\z@}%
  {0pt}
  {0pt}%                % No space after the heading
  {\normalfont\bfseries}% Bold, normal (non-italic) text
}
\makeatother

\let\oldparagraph\paragraph
\renewcommand{\paragraph}[1]{\oldparagraph{#1}: }

\begin{document}
\title{Reinforcement Learning for AMR Charging Decisions: The Impact of Reward and Action Space Design}
\titlerunning{RL for AMR Charging Decisions}
% If the paper title is too long for the running head, you can set
% an abbreviated paper title here
%
\author{Janik Bischoff\inst{1}\orcidID{0009-0007-6592-9768} \and
Alexandru Rinciog\inst{2}\orcidID{0000-0003-0330-6737} \and
Anne Meyer\inst{1}\orcidID{0000-0001-6380-1348}}
\authorrunning{Bischoff et al.}
% First names are abbreviated in the running head.
% If there are more than two authors, 'et al.' is used.
%
\institute{Karlsruher Institut für Technologie, Zirkel 2, 76131 Karlsruhe \and
SLAPStack, Joseph-von-Fraunhofer-Straße 2-4, 44227 Dortmund
\email{alexandru.rinciog@slapstack.de}\\
% \url{http://www.springer.com/gp/computer-science/lncs}
\email{\{janik.bischoff,anne.meyer\}@kit.edu}}
\maketitle              % typeset the header of the contribution
%
% \small
\input{content/0_abstract}
\setcounter{tocdepth}{3}
%
%
%
\input{content/1_introduction}
\input{content/2_literature}

\input{content/3_problem}
\input{content/4_rl4bm}
\input{content/5_experiments}

\input{content/6_conclusion}
\newpage
%
% ---- Bibliography ----
%
% BibTeX users should specify bibliography style 'splncs04'.
% References will then be sorted and formatted in the correct style.
%
% ---- Bibliography ----
\bibliographystyle{splncs04}
\bibliography{wepa_bm}
\end{document}

%% file: content/0_abstract.tex
\begin{abstract}
We propose a novel reinforcement learning (RL) design to optimize the charging strategy for autonomous mobile robots in large-scale block stacking warehouses.
RL design involves a wide array of choices that can mostly only be evaluated through lengthy experimentation. 
Our study focuses on how different reward and action space configurations, ranging from flexible setups to more guided, domain-informed design configurations, affect the agent performance. 
Using heuristic charging strategies as a baseline, we demonstrate the superiority of flexible, RL-based approaches in terms of service times.
Furthermore, our findings highlight a trade-off: While more open-ended designs are able to discover well-performing strategies on their own, they may require longer convergence times and are less stable, whereas guided configurations lead to a more stable learning process but display a more limited generalization potential.  
Our contributions are threefold. 
First, we extend SLAPStack, an open-source, RL-compatible simulation-frame\-work to accommodate charging strategies. 
Second, we introduce a novel RL design for tackling the charging strategy problem.
Finally, we introduce several novel adaptive baseline heuristics and reproducibly evaluate the design using a Proximal Policy Optimization agent and varying different design configurations, with a focus on reward.

\keywords{Autonomous Block Stacking Warehouses  \and Vehicle Dispatching \and Discrete Event Simulation \and Battery Management \and Reinforcement Learning \and AGV}
\end{abstract} 

%% file: content/1_introduction.tex
% =========================================================================================
\section{Introduction}
\label{sec:introduction}
%==========================================================================================  
With increasing uncertainty and supply chain disruptions, innovative logistics solutions are becoming vital. The adoption of autonomous mobile robots (AMR) in block storage warehouses (BSW) reflects this trend.

% problem in general
In a BSW, goods are stored directly on the floor or on top of each other. 
This storage system has several benefits, such as low investments due to the minimal infrastructure needed and high throughput.  
While pallets are stored in lanes next to each other for higher storage efficiency, aisles and cross-aisles are used by vehicles to travel through the BSW. 

The authors of \cite{pfrommer_autonomously_2020} have presented the autonomous block stacking warehouse problem (ABSWP), which consists of interdependent decision problems in combination with the usage of AMRs in a BSW.
Problems include the storage location assignment problem (SLAP), which assigns incoming items to storage locations, the unit load selection problem (ULSP) to determine which items are retrieved to fulfill an outbound order, and the vehicle dispatching problem, in which vehicles are assigned to transport tasks. Due to the combinatorial complexity of these decision problems, simulation studies are used to derive configurations for a given ABSWP instance.
The importance of battery management is often overlooked in such studies.
% the crucial role that effective battery management plays, as highlighted by \cite{McHANEY1995} and \cite{Vis2006}. 
Recognizing the crucial role that battery management plays in autonomous systems, as highlighted in  \cite{McHANEY1995,Vis2006}, we extend the ABWSP model to include it. 

\begin{figure}[ht]
    \vspace{-10pt}
    \centering
    \includegraphics[width=0.95\textwidth]{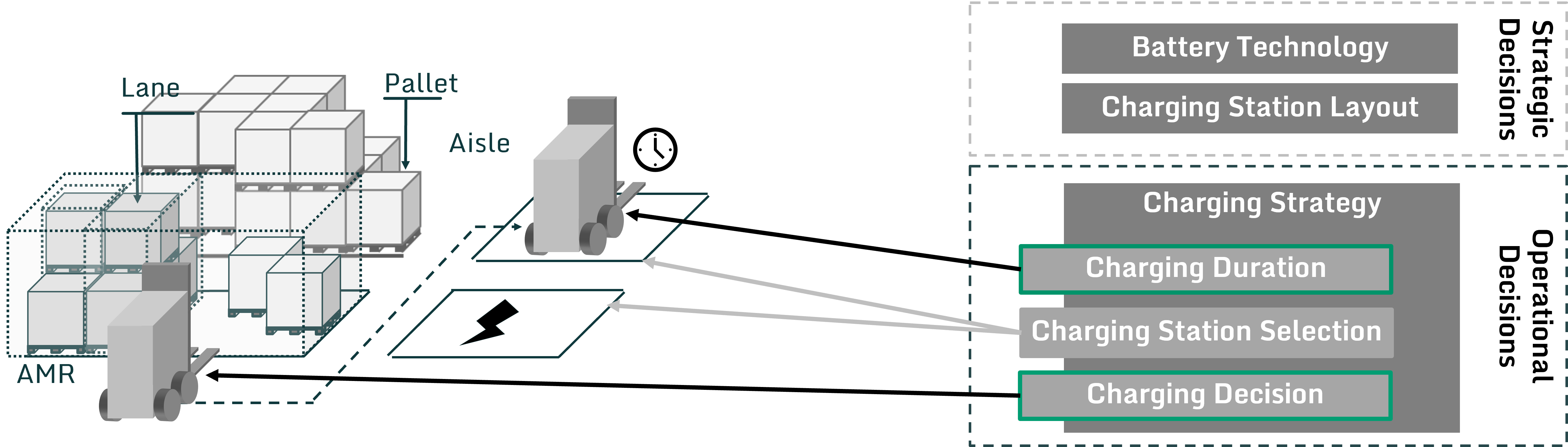}
    \caption[Battery Management Problem]{\small The problems associated with battery management as per  \cite{Wang2014}.}
    \label{fig:battery_management_problem}
    \vspace{-10pt}
\end{figure} 

%ABSATZ
% charging dispatching
According to \cite{Wang2014}, battery management consists of three sub-problems visualized in Figure \ref{fig:battery_management_problem}: The selection of battery charging technology (e.g. induction charging, battery swaps), the charging station layout (i.e. the number and position of charging stations), and the battery charging strategy. 
The former problems are more strategic in nature, while the last problem is operational.
%We focus on the operational part of battery management, the selection of a suitable battery charging strategy.
The charging strategy, which is the focus of this work, can be further broken down into three sub-problems pertaining to (1) the \textbf{charging decision}, i.e. determining when an AMR should go charge, (2) the \textbf{charging station selection}, i.e. assigning a specific charging station and the route to that destination to an AMR, and (3) the \textbf{charging duration}, i.e. determining how long AMRs should remain at charging stations.

%ABSATZ
% motivate RL
% frontloading: this work focuses on an RL meta-heuristic solution 
This work employs a reinforcement-learning (RL) meta-heuristic approach to jointly solve the charging decision and charge duration problems. 
To accomplish this, the agent must decide between discrete charge levels including a no-charge option, at the conclusion of any travel event.
We use a fixed heuristic solution for the charging station selection problem and evaluate the resulting charging strategy using a large-scale, real-world ABSWP benchmark --- WEPAStacks \cite{Pfrommer2022}.
%because...
Charging strategy solutions can be either exact, heuristic, or meta-heuristic. 
Exact solutions are intractable for large real-world scenarios such as WEPAStacks.
Simple heuristic solutions are likely not able to adapt to varying AMR demand, resulting in inefficiency.
Meta-heuristic solutions provide more adaptability %than simple heuristics 
while maintaining tractability.
We chose RL as the meta-heuristic paradigm for two reasons.
Firstly, in recent years, RL has emerged as a promising approach for solving complex combinatorial optimization problems \cite{Kuhnle2019}.
% reason 1: Promising results in related fields
%Following its success in domains such as gaming or robotics, reinforcement learning (RL) has emerged as a promising approach to solving complex combinatorial optimization problems \cite{Kuhnle2019}.
% reason 2: Prior work -> simulation already available
%As we can leverage existing simulation environments for the ABSWP we can employ RL to learn battery management decisions in the context of this simulation.
Secondly, an RL-compatible ABSWP simulation with mechanisms safeguarding reproducibility was made available in \cite{Pfrommer2022}, alongside real-world use-case data.
This makes the RL training and evaluation effort manageable. 
Among various RL algorithms, we selected Proximal Policy Optimization (PPO) \cite{schulman2017proximal} due to its suitability for practical applications in combinatorial optimization \cite{mazyavkina2021reinforcement}.

% GOAL & contributions
A key challenge in applying RL lies in the multitude of design choices such as the design of the reward function and action space. 
We explore different configurations varying the amount of structure imposed on the agent’s decision-making process.  
Our results suggest a tradeoff between learning stability and optimization potential.
A more flexible approach to RL-design provides greater optimization potential at the cost of learning stability.
Conversely, a more constrained, domain-knowledge informed design, can lead to good results quickly, but may limit the agent's learning potential.
% this work> contribution
Overall, we make the following contributions to the field:
\begin{enumerate}
    \item Firstly, we \textbf{extend SLAPStack}, a fine-grained, rl-compatible, ABSWP simulation framework \cite{Pfrommer2022}, to account for battery management, and publish the associated code.
    \item Secondly, we introduce a \textbf{novel Markov decision process (MDP)} for AMR charging.
    \item Finally, we train a state-of-the-art RL algorithm on a large-scale ABSWP instance and \textbf{reproducibly evaluate several different MDP-con\-fi\-gu\-ra\-tions} against existing and \textbf{novel heuristic strategies}. 
\end{enumerate}
To the best of our knowledge, this is the first work to model battery-aware AMR behavior using deep RL in a realistic, large-scale warehouse environment in a reproducible fashion.

We start with an overview of %RL-based and learning-free 
battery charging strategies in literature in Section \ref{sec:lit}. 
We then present the problem setting and benchmark instance in Section \ref{sec:problem-setting}, including a selection and evaluation of baseline charging strategies. % and evaluate them.
In Section \ref{sec:rl4bm}, we present the different RL design choices for AMR charging, and detail the configurations we evaluate. 
Finally, we evaluate the proposed configurations in terms of learning stability and generalization capabilities in Section \ref{sec:experiments}, before drawing our conclusions in Section \ref{sec:conclusion}.

%% file: content/2_literature.tex
%==========================================================================================
\section{Related Work}
\label{sec:lit}
While some research exists on battery management for AMRs in intralogistics, no existing work offers suitable charging strategies for real-world ABSWPs. Our review shows that exact methods handle only small problems, heuristics lack flexibility, and RL approaches suffer from non-reproducible simulations.
\label{sec:lit-cdp}

\paragraph{Learning Free Approaches}
The lower level decisions comprising battery charging strategies, i.e. deciding when to charge, for how long, and which charging station to use, are often tackled using simple heuristic rules or expert knowledge. 
%A common strategy is to parameterize the decision of when to charge using fixed thresholds, for example 20 \% to prevent the AMR from blocking production due to insufficient battery levels \cite{9928864,https://doi.org/10.2195/ljproccollingen20191201}. 
A common approach for deciding when AMRs should charge is to use fixed lower-bound battery capacity thresholds.
In \cite{9928864,https://doi.org/10.2195/ljproccollingen20191201}, for instance, the 20\% level is chosen to prevent AMRs from blocking production due to insufficient battery levels. 
A more flexible approach is opportunity charging. 
Here, charging is done when an opportunity arises. 
This may occur when vehicles are idle and near a charging station or during off-peak times. 
For this strategy, it is important that the placement of the charging station coincides with the regions where AMRs are likely to idle, e.g. on the way to a central parking area or near an in- or outbound dock as described in \cite{ca9569745e1349e5a93f049189306a32}. 

While modern AMRs are able to partially recharge via inductive or plug-in charging, battery swapping is still widespread. 
Herein, the empty battery is replaced with a full one at designated swapping stations. 
Although battery swapping eliminates idle time for the AMR during charging, it introduces significant disadvantages. 
It is less safe than charge-based techniques and requires special safety measures such as acid-proof floors at the swapping stations, as noted in \cite{ca9569745e1349e5a93f049189306a32}. 
% Further, the additional complexity of constructing swapping stations requires a higher investment, thus increasing the fixed costs and capital expenditure for such a system. 
For inductive or plug-in charging systems, a common practice is to recharge to a defined upper level, e.g. 50\% / 80\% / 100\% as in \cite{McHANEY1995,https://doi.org/10.2195/ljproccollingen20191201,Zhan2019,Kabir2018}.  
The charging station selection is also often done in a rule-based fashion, e.g. selecting the closest or the least queued charging station \cite{ca9569745e1349e5a93f049189306a32,deKoster2007}.

In addition to these myopic strategies, mixed integer linear programming (MILP) is sometimes applied to determine optimal charging schedules for AMR dispatching as in \cite{Singh2022,Meyer2024}. In both works, the computational efforts to solve larger instances are mentioned. To overcome this Meyer et al formulate a branch-price-and-cut approach which can solve instances with up to 144 tasks in \cite{Meyer2024}. They model the problem of assigning transport tasks in intralogistic warehouses to AMRs as an electric vehicle routing problem. Two objectives are proposed to minimize completion times and due dates, respectively. The authors compare battery swapping, partial charging, and full recharging strategies. 
Besides a MILP program the authors of \cite{Singh2022} present a matheuristic approach to solve instances larger than 22 tasks. 
% constraints is presented. They solved instances with up to 22 tasks optimally. To solve larger instances, a matheuristic approach that is a combination of an adaptive large neighborhood search and linear programming is presented. 
AMRs can partially recharge after a critical battery threshold is reached. For recharging, the nearest charging station is selected.

Metaheuristic approaches have been proposed in several other works. 
Mousavi et al. employ a hybrid genetic algorithm-particle swarm algorithm to optimize an AMR scheduling process in a flexible manufacturing system \cite{Mousavi2017}. A constraint ensures that the charge level of an AMR is sufficient to fulfill a given transport order. 
The objective function minimizes both the makespan and number of AMRs required for the transport tasks. The problem sizes range from 6 to 15 transport jobs.
Han et al. use a genetic algorithm to solve an AMR scheduling problem in \cite{Han2022}. A constraint prevents AMRs from accepting transportation orders if their battery level is below 20 \%. First, a regular scheduling is carried out, then the schedule is repaired to account for necessary charging tasks. The approach is applied to a case study with 29 tasks.

From the problem sizes used in these works, it is evident that MILP and other exact methods are ill-suited for dynamic, online variants of the ABSWP, which require frequent re-optimization in response to new tasks. 
This is due to the strong interdependence between battery management and other sub-problems, meaning that storage assignment, routing, and task precedence must be decided jointly with charging-related variables. 
%As a result, exact methods are currently inapplicable to realistic problem sizes involving thousands of storage locations and hundreds of vehicles and orders. 
Proposed Meta-heuristics other than RL also exist, but these are also typically designed for offline settings with limited task volumes.
\paragraph{Reinforcement Learning Approaches}
In \cite{Lin2023} a feature-based SARSA algorithm is formulated to determine the charging duration for AMRs. AMRs below a certain threshold are required to go to a standby area. If then a free charging station is available, the AMR occupies it. The action is to determine the charging duration of each AMR at the charging station according to the system state and the current energy price.
The RL approach is compared to an industry heuristic that uses a safety threshold of 20 \% and aims to keep the battery level between 40\% and 80\%. 
The observation space consists of information on the AMRs in the standby area and at the charging stations, such as battery charge levels, battery ages, and battery types of the AMRs. 
50 AMRs are considered with 0.8 transportation tasks arriving each minute over a course of 30 days. Charging decisions are made every time a task arrives. The authors report improved utilization for their approach.

The authors of \cite{Deng2020,Mu2022} present an RL approach for an automated warehouse setting. Twin Delayed Deep Deterministic Policy Gradient is used as the RL algorithm.
At each time step, a set of AMRs with the lowest battery in the working area has to go charging, and charging AMRs with a battery level above a certain threshold must return to the working area. The action is to select the number of AMRs to go charging and the working battery threshold.
A setting with 640 AMRs and 3,500 to 4,500 orders per day is used to train and evaluate the approach. They simulate 25 days and divide each day into 510 time steps. The number of orders fulfilled per step is used as the reward function. Their results are compared with rule-based charging strategies that are parameterized by a working and charging threshold that determine when the AMRs have to charge and when they have to return to the working area. 

In \cite{Singh2024} an RL is applied to the dispatching of heterogeneous AMR in an online setting. A Noisy Dueling Double Deep Q-Network which enhances exploration via NoisyNets is employed. Their approach is trained on instances containing up to 12 AMRs and 900 tasks. The approach is compared to the exact and matheuristic methods in \cite{Singh2022}.

% Gap
From the reviewed RL approaches we identify a number of gaps. 
In addition to the missing publication of the simulation and possibilities to reproduce the results, neither of the proposed approaches evaluates different RL design spaces but present single action and reward spaces. 
The approaches also differ in terms of decision granularity. In \cite{Lin2023} decisions are made at the level of individual AMRs, but only the charging duration is optimized. 
The decision of when to go charging is fixed. 
In \cite{Deng2020,Mu2022} both the number of AMRs to go charging and a target charging threshold are determined, but on an aggregated level without taking the states of the individual AMRs into account.

%% file: content/3_problem.tex
%==========================================================================================
\section{Benchmark Extensions and Baseline Evaluation}
\label{sec:problem-setting}
Meaningful RL experiments require a strong grasp of the experimentation framework and its underlying data. 
This section outlines the battery management extensions in SLAPStack and WEPAStacks, and compares the heuristic charging strategies used as RL baselines.
To keep within the frame of this work, we defer the implementation details to the published repository \cite{bischoff2025slapstack} and elaborate only on the parts needed to understand the RL design discussion that follows.

\subsection{Benchmark Framework Extension}\label{subsec:extensions}
\paragraph{Augmented SLAPStack}
%% SLAPStack HIGH LEVEL DETAILS
SLAPStack is a discrete event simulation implementing an intuitive event chain concept detailed in its public GitHub repository \cite{rinciog2023slapstack}. 
Event chains model the process logic of the block storage and can pause the simulation execution to request inputs from control algorithms.
The delivery event chain, for instance, is implemented as follows: When a delivery order arrives, an AMR is sent to the dock to pick up the pallet.
When it arrives at the dock, the simulation requests a decision from a SLAPStrategy. 
The vehicle then moves the pallet to the indicated position, where it is released upon arriving.

Following this pattern, we implemented a charging event chain that requires a decision after every completed transportation task.
An external decision entity determines how long an AMR should charge. 
A value of 0 indicates no charging, while any other value up to 100 specifies both the decision to charge and the desired duration. 
The vehicle then goes to the charging station and remains there for a corresponding time period, after which the vehicle is released.
On every AMR movement, its battery is depleted based on movement time and load.
The capacity replenishment occurs on vehicle release at a charging station.
%% Battery Management Event chain Extension

%% BATTERY MODEL
We assume vehicles that are capable of automated plug-in or inductive charging. 
Following the assumptions of \cite{McHANEY1995} the battery capacity of the vehicles is set to 52 ampere hours (Ah). We also use the same consumption rates of $travel_{\text{loaded}} = 15 Ah$ and $travel_{\text{unloaded}} = 10 Ah$.
Further, a charging duration of half an hour to recharge to 100\% is assumed.
For recharging the battery, we assume a linear charging behavior as no non-linear movement model was available for this work. 
Battery degradation effects are not considered in this study.

\paragraph{Augmented WEPAStacks}
Each SLAPStack use-case is defined by three components: warehouse layout, order set and an initial fill level. 
The initial fill level maps different stock keeping units (SKU) to the volume present in the warehouse. 
The order stream defines the times at which pallets with associated SKUs need to be stored or retrieved from the warehouse along with their in-/output point (dock).
The layout defines the spatial dimension of all warehouse elements, i.e. storage, aisles and docks.
Battery management augments this setup by specifying charging station locations.
 
\begin{figure}
    \vspace{-10pt}
    \centering
    \subfloat[\small Warehouse layout.]{
        \includegraphics[width=0.38\linewidth]{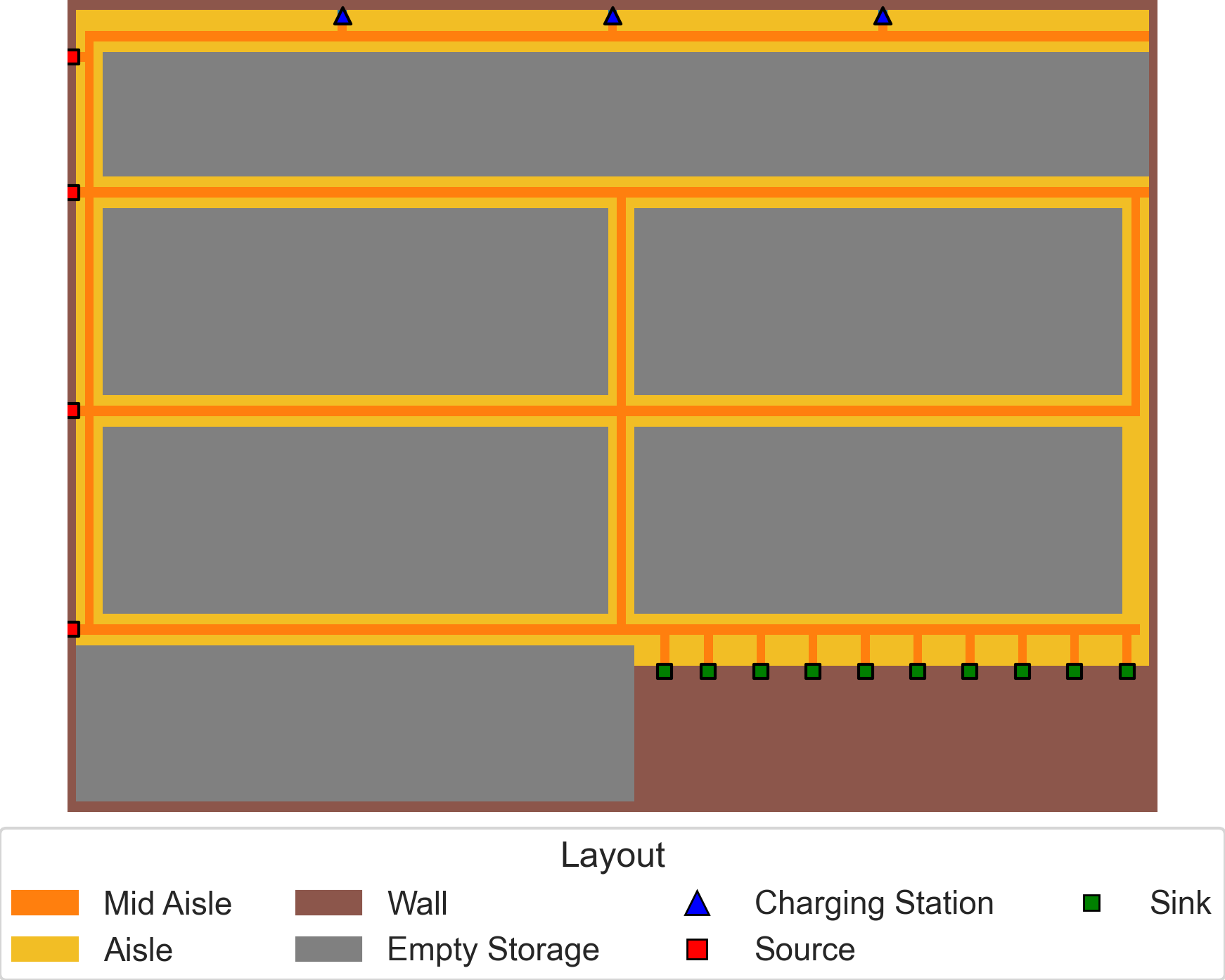}
        \label{subfig:layout}
    }
    \hfill
    \subfloat[\small Order arrival distributions per week.]{
        \includegraphics[width=0.55\linewidth]{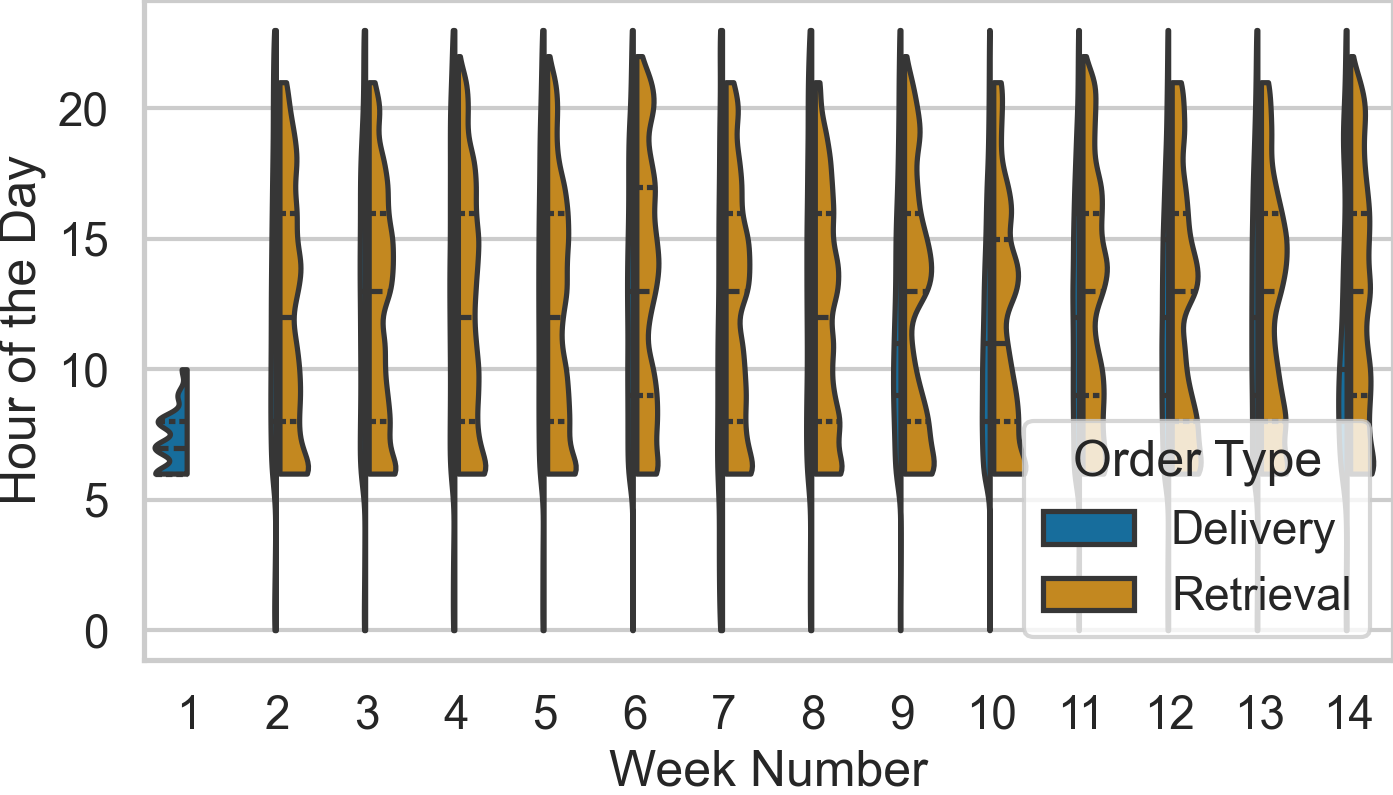}
        \label{subfig:orders}
    }
    \caption{\small WEPAStacks use-case data visualization.}
    \label{fig:layout}
    \vspace{-10pt}
\end{figure}

WEPAStacks models a finished goods warehouse located at the WEPA GbmH hygiene paper company production site. 
Figure \ref{fig:layout} visualizes the augmented layout and the order stream of the dataset we employ.
Three charging stations are placed equidistantly along the north warehouse wall  (Figure \ref{subfig:layout}). 
The warehouse is 150 by 80 meters, and has 4 input and 10 output docks.
The order stream contains 400,000 in- and outbound orders spanning 89 days.
Figure \ref{subfig:orders} displays the order distribution over hours aggregated on a weekly basis.
Outbound trucks arrive from 06:00 am to 22:00 pm, while the production outputs are continuous.

To prevent truck and, in particular, production queues in WEPAStacks, the number of pending retrieval and delivery orders is limited to 330 and 240, respectively, as described in \cite{Pfrommer2022}.  
We determine the required number of charging stations using a workflow similar to the one used in \cite{Rinciog2023}, where the nr of AMRs is inferred.
We first fix simple heuristic solutions for all battery-augmented ABSWP decisions.
Then, starting from a single charging station, we run simulations while gradually increasing their numbers until all system constraints are satisfied. 

We use the solvers for the ABSW subproblems as reported in \cite{Pfrommer2022}. 
For SLAP we employ the closest open pure lanes strategy. 
Here, SKUs are stored in the closest lane that contains only that SKU. 
If no such lane exists, the nearest open location is chosen.
The ULSP is solved using a last in first out strategy, in which the SKU that arrived last at the warehouse is retrieved first. 
For vehicle dispatching, the nearest available vehicle is selected. 
%%%%%%%%%%%%%%%%%%%%%%%%%%%%%%%%%%%%%%%%%%%
\subsection{Baseline Strategies}
\paragraph{Definition} We introduce three strategies to address the charging-decision and charging-duration problems.
\begin{itemize}
    \item \textit{Fixed-threshold}: A fixed-threshold strategy is parameterized by two thresholds: $TH_{lower}$, which determines the battery level at which an AMR has to go charge, and $TH_{upper}$, which specifies the target battery level. 
    We set $TH_{lower}$ to 20 \% and vary $TH_{upper}$ from 30 \% to 100 \% in steps of ten.
    \item \textit{Opportunity}: The opportunity-charge definition from \cite{ca9569745e1349e5a93f049189306a32}, which was explained in Section \ref{sec:lit}, is not applicable in our scenario since the charging stations are not placed near the inbound and outbound docks.
    In our context, opportunity charging refers to charging when stations are available, and no work is pending. 
    $TH_{upper}$ is fixed to 100\%. 
    \item \textit{HighLow}: In the high-low approach, $TH_{lower}$ is fixed while $TH_{upper}$ varies depending on the retrieval order queue length. 
    If there are queued retrieval orders, we use $TH_{upper}$; otherwise the battery is fully re-charged. 
    To determine a good value for $TH_{lower}$, we tune it using the same thresholds as in the \textit{Fixed-threshold} strategy.
\end{itemize}

The authors of \cite{https://doi.org/10.2195/ljproccollingen20191201} highlight that non-adaptive charging strategies often result in too few AMRs being available to handle tasks. 
To alleviate this, we propose the following improvement:
\begin{itemize}
    \item \textit{Interrupt}: If retrieval orders are pending and no AMR is available, the interrupt strategy halts charging for any AMR whose battery level exceeds the upper threshold $TH_{interrupt}$. 
    In this work, we set $TH_{interrupt}$ to 50 \% 
\end{itemize}

For all strategies, we fix the selection of the charging station: the nearest available charging station is prioritized; if none is free, the station with the shortest queue is chosen. 

\begin{figure}
    %\vspace{-10pt}
    \centering
    \includegraphics[width=\linewidth]{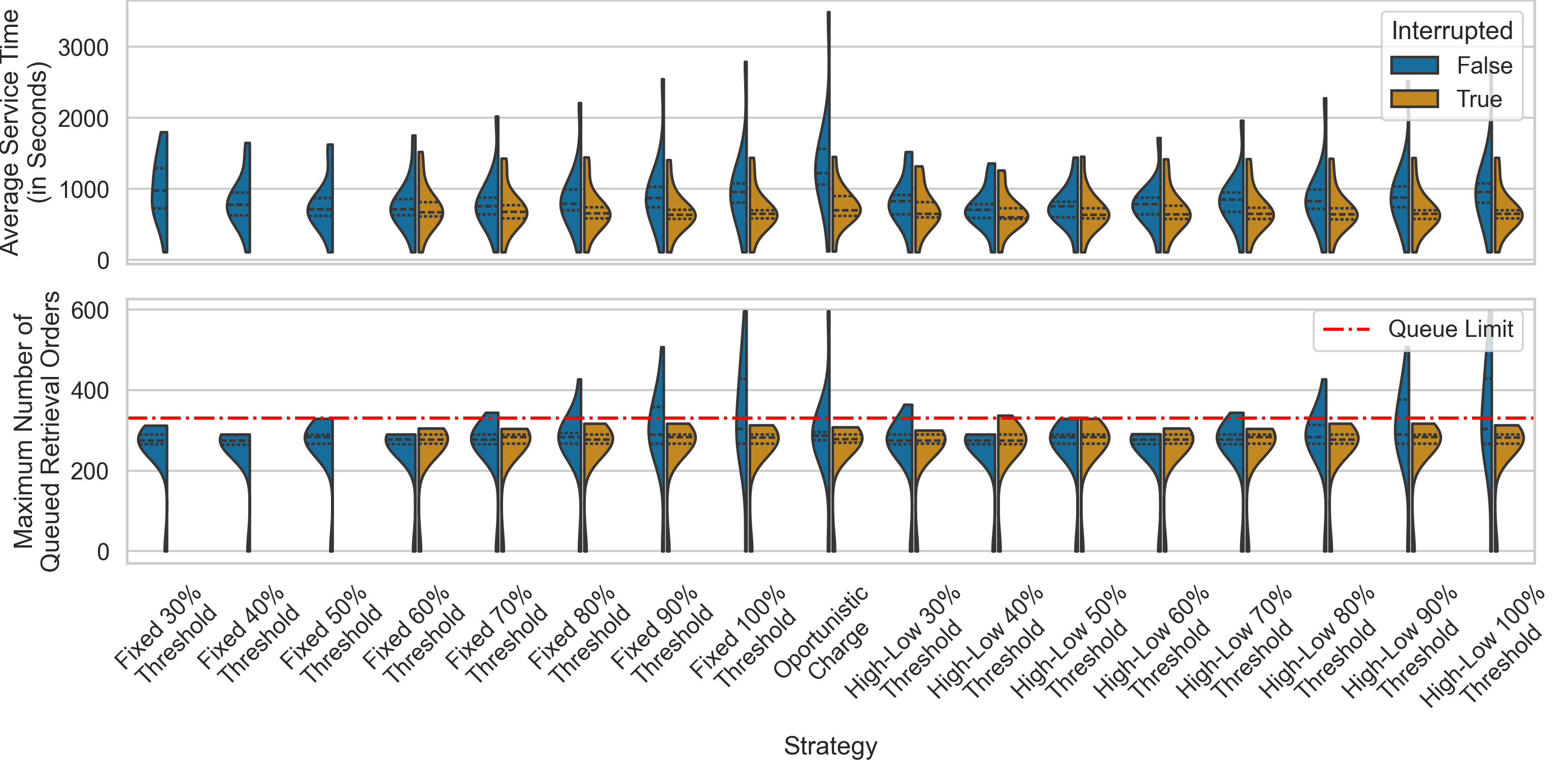}
    \caption{\small Week and intercept dependent distribution of service times and buffer sizes.}
    \label{fig:baseline_results}
    \vspace{-10pt}
\end{figure}
\paragraph{Evaluation}
In Figure \ref{fig:baseline_results}, we show the results of the proposed baseline strategies for all weeks as violin plots. 
We compare the strategies in terms of average service time and maximum number of queued retrieval orders, both with and without the interrupt scheme. 
The figure suggests that 
%We can see that successful battery 
charging strategies can help maintain AMR availability, either through short charging bursts, as \textit{Fixed-threshold} with $TH_{upper}=40\%$, or by interrupting the charging process. 
\textit{Fixed-threshold} strategies with target thresholds of 80\% to 100\% violate queue constraints and result in long service times. 
This is in line with the results of \cite{Meyer2024}, where the poor performance of full recharge strategies was highlighted. 
Implementing the interrupt scheme consistently improves all metrics, with the most significant improvements observed in high-threshold configurations. 
For example, applying interruption to the \textit{Fixed-threshold} strategy with $TH_{upper}=100\%$ shows a 33\% reduction in service time and 47\% reduction in maximum queue size. 
% The strategy \textit{HighLow} that varies short/long charging cycles depending on the system load, leads to good results in both average service time and queued retrieval orders. 
Notably, the \textit{HighLow} strategy with $TH_{upper}=40\%$ achieves the best overall performance among non-interrupted strategies, with an average service time of 741 seconds and maximum queue size of 290 orders. 
This suggests that adapting charging durations based on the system load helps to maintain AMR availability.
With interruption enabled, the benefits of the \textit{HighLow} strategy are reduced. 
In this case, \textit{HighLow} with $TH_{upper}=90\%$ performs best, closely matching the corresponding \textit{Fixed-threshold} strategy. 

The \textit{Opportunity} strategy on the other hand is not able to satisfy the constraints. 
In addition, it leads to the highest average service times. 
This indicates flaws in the design of the heuristic.

%% file: content/4_rl4bm.tex
\section{A Reinforcement Learning Charging Strategy}
\label{sec:rl4bm}
In this section, we outline the proposed approach to AMR charging using RL. We present the core components of the MDP, and motivate the use of PPO. 
% such as the reward function, state space, and action space
Following the extension described in Section \ref{subsec:extensions}, SLAPStack now supports training and deploying RL agents for charging decisions in addition to SLAP, and ULSP.
Since this work focuses on charging decisions, we adopt the best-performing heuristic solvers from \cite{Pfrommer2022} to handle SLAP and ULSP in the background. 
The simulation only pauses to request a charging decision from the RL agent.
A charging decision is triggered after any completed transport task.% when an AMR has no active transport task.

\paragraph{State Representation and Feature Design} 
Using the full system state of the ABSWP --- encompassing all vehicle movements and storage locations --- as the agent input would be computationally prohibitive. 
In line with \cite{Mu2022} and \cite{Deng2020}, we adopt a feature-based state representation. 

This includes AMR-related and operational features, such as battery level, distance to the charging stations, and the number of pending charging events and orders. 
We further incorporate ABSWP-specific features, most notably the average lane-wise entropy introduced in \cite{Pfrommer2022}, which takes a value in $[0, \infty)$, with lower values indicating more ordered lanes. 

As shown in \cite{Raichuk2021WhatMF}, normalizing the observation space is essential for on-policy deep actor-critic algorithms such as PPO. 
We therefore scale all features to the $[0, 1]$ range.
Table \ref{tab:rlfeatures} provides the formal definitions of these features using the notation in Table \ref{tab:notation}.

\begin{table}[h]
    \vspace{-10pt}
    \caption{\small Feature space variables.}
    \centering
    \scriptsize
    \begin{tabularx}{\linewidth}{p{0.35\linewidth}X}
        \toprule
        \textbf{Symbol} & \textbf{Description} \\
        \midrule
        $\mathcal{V}, \mathcal{C}$ & Set of AMRs ($\mathcal{V}$) and charging station ($\mathcal{C}$) indices \\
        $\mathcal{L}, \mathcal{O}$ & Set of lane ($\mathcal{L}$), and finished order ($\mathcal{O}$) indices \\
        $B$ & Maximum battery capacity \\
        $B^{\text{cs}}_n$ & Battery level of AMR at charging station $n$ \\
        $V^{\text{bl}}_j$ & Battery level of AMR $j$ \\
        $V^{\text{depleted}}_j$ & 1 if AMR $j$ is depleted; 0 otherwise \\
        $V^{\text{busy}}_j$ & 1 if AMR $j$ is busy; 0 otherwise \\
        $V^{\text{free}}_j$ & 1 if AMR $j$ is free; 0 otherwise \\
        % $V^{\text{amr}}_n$ & Battery level of the AMR at charging station $n$ \\
        $q_n$ & Queue length at charging station $n$ \\
        $q_r, q_d$ & Queue lengths of retrieval and delivery orders \\
        $q^{\text{max}}_r, q^{\text{max}}_d$ & Maximum queue length for retrieval and delivery orders \\
        $T^{\text{ol}}, T^{\text{nl}}$ & Number of occupied and total storage locations \\
        $h, d$ & Current hour and current day \\
        $TR_i, DR_i$ & Travel time ($TR_i$), and distance ($DR_i$) for retrieval order $i$ \\
        $TD_i, DD_i$ & Travel time ($TD_i$), and distance ($DD_i$) for delivery order $i$ \\
        $L_k$ & Warehouse Lane $k$\\
        $p_{\text{SKU}}$ & Relative SKU amount in a Lane \\
        \bottomrule
    \end{tabularx}
    \label{tab:notation}
    \vspace{-10pt}
\end{table}

\begin{table}[htbp]
\caption[Reinforcement Learning features]{\small Features used during RL-training.}
\centering
\scriptsize
\begin{tabularx}{\textwidth}{>{\hsize=0.08\hsize}X >{\hsize=0.55\hsize}X >{\hsize=0.37\hsize}X}
  \toprule
  \textbf{F\textsubscript{i}} & \textbf{Feature Name} & \textbf{Definition} \\
  \midrule
  1 & Mean battery level for all AMRs & $\frac{1}{|\mathcal{V}|} \sum_{j\in\mathcal{V}} \frac{V^{\text{bl}}_j}{B}$ \\ 
  2 & Mean battery level for all busy AMRs & $\frac{1}{\sum_{j\in\mathcal{V}} V^{\text{busy}}_j} \sum_{j\in\mathcal{V}} \frac{V^{\text{bl}}_j \cdot V^{\text{busy}}_j}{B}$ \\
  3 & Battery level of charging AMRs & $B^{\text{cs}}_n, n \in \mathcal{C}$ \\
  4 & Battery level of AMRs & $V^{\text{bl}}_j, j \in \mathcal{V}$ \\
  \midrule
  5, 6, 7 & Number of currently depleted/free/busy AMRs & $\frac{1}{|\mathcal{V}|} \sum_{j\in\mathcal{V}}V^{\text{depleted | free | busy}}_j$ \\
  8 & Overall fleet utilization & $\frac{1}{|\mathcal{V}|} \sum_{j\in\mathcal{V}} V^{\text{busy}}_j$ \\
  \midrule
  9 & Warehouse fill level & $1 - \frac{T^{\text{ol}}}{T^{\text{nl}}}$ \\
  10 & Number of queued AMRs at charging stations & $q_n, n \in \mathcal{C}$ \\
  11 & Number of queued retrieval, and delivery orders & $\frac{q_r}{q^{\max}_r}$, $\frac{q_d}{q^{\max}_d}$ \\
  \midrule
  12 & Hour (sinusoidal encoding) & $\sin(2\pi \cdot h \cdot 24^{-1})$, $\cos(2\pi \cdot h \cdot 24^{-1})$ \\
  13 & Day of the week & $d \in \{1,\ldots,7\}$ \\
  \midrule
  14 & Number of currently free charging stations & $\left| \{n \in \mathcal{C} \mid q_n = 0 \} \right|$ \\
  15 & Average lane-wise entropy & $- \frac{1}{\vert \mathcal{L} \vert}\sum_{k \in \mathcal{L}} \sum_{\text{SKU} \in L_k} p_{\text{SKU}} \log(p_{\text{SKU}})$ \\
  16, 17 & Average travel time retrieval | delivery & $\frac{1}{|\mathcal{O}|} \sum_{i\in\mathcal{O}} TR_i|TD_i$ \\
  18, 19 & Average distance retrieval | delivery & $\frac{1}{|\mathcal{O}|} \sum_{i\in\mathcal{O}} DR_i|DD_i$\\
  \bottomrule
\end{tabularx}
\label{tab:rlfeatures}
\vspace{-10pt}
\end{table}

\paragraph{Reward and Action Space Design}
% Explain RL design configurations
We propose several reward and action space configurations that progressively transition from a broad, flexible design to a more structured one. 
The underlying intuition is that a wider action space and sparse rewards offer more room for the agent to discover novel strategies, while a narrower, reward-engineered approach --- augmented by domain knowledge --- may lead to faster convergence, but could limit the discovery of high-performing strategies.

We vary configurations in terms of the reward function definition, action space, and use of the interrupt heuristic. 
% Present and motivate action space
We utilize a discrete action space $A$. 
At each time step $t$, given agent state $S_t$ the agent can select a non-zero charging time threshold, or decide not to charge by selecting the $0$ action. 
%Not charging is encoded as action $A_0 = 0$. 
This allows us to address both charging decision and charging duration sub-problems. 
The number of target thresholds is a design choice. 
In this work, we employ two alternatives.
$A_{full}:=[0, 30, 40, 50, 60, 70, 80, 90, 100]$ contains all the thresholds used in our baseline evaluation. 
$A_{binary} :=[0,100]$ is a binary action space, triggering full charge or no charge at all. 

Using discrete charging thresholds instead of continuous durations, we enable action masking, which prevents agents from taking illegal actions.
Action masking applies a binary mask over the action space at time step $t$ to mask out invalid actions by setting their sample probability to negative infinity \cite{Huang2022}. 
Action masking is crucial for our application to prevent infeasible battery states, e.g. a target battery level lower than the current one. 
We apply the following action mask to determine feasible actions for an AMR at time step $t$: 
Action $A_i$ $\in A$ is feasible if $A_i > V_t^{bl}$, where $V_t^{bl}$ denotes the battery level of the AMR being considered at time step $t$.
This allows us to tackle both charging decisions and charging duration sub-problems.
We use 20\% as the final battery level where recharging is required to prevent AMRs from stranding during operation.  
Therefore, if $V_t^{\text{bl}}<=20$, action 0 becomes invalid and charging becomes mandatory.

% Motivate reward
To the best of our knowledge, there is no universal approach for constructing a reward function. 
The design of an effective reward function involves several challenges, such as managing competing objectives.
The so-called credit assignment problem \cite{Sutton1998} is another major challenge: Due to the delayed nature of rewards, it is difficult to identify the actions that led to a particular outcome.
For these reasons, finding a suitable reward is often a trial-and-error process \cite{Booth2023}. 
To overcome the credit assignment problem, reward shaping is often used, where domain and expert knowledge are incorporated to guide the agent's learning process, thus providing more direct feedback \cite{Sutton1998}.

In the context of the ABSWP, our main objective is to minimize the average service time $ST_{avg}$, defined as the sum of all service times divided by the number of completed orders. 
The service time $ST_i$ for an order $i \in \mathcal{O}$ is the difference between its completion time and arrival time. 
In addition, we want to incentivize the agent to stay within the order queue limits. % as described earlier. 
To achieve these goals, we propose four reward functions that progressively incorporate higher levels of guidance: % and domain knowledge:
\begin{itemize}
    \item \textbf{Service time-based reward:} The first reward function directly penalizes the agent based on the negative average service time observed at the current decision step $-ST_{avg}$, thus aiming to directly optimize for the main objective.

    \item \textbf{Queue-based reward:} The second reward function is $-q_r + q_d$, representing the negative sum of all queued retrieval and delivery orders in the current step. 
    Although it does not explicitly optimize service time, our baseline analysis shows that high queue levels can lead to increased service times.  

    \item \textbf{Composite reward + free AMR component:} %As shown in the previous section, good charging strategies are able to provide AMRs when needed. Therefore, 
    This reward function combines the previous two components, while introducing an additional incentive to keep AMRs available during busy system states:  
    when orders are pending, the agent receives a bonus for having free AMRs available, as this increases flexibility in fulfilling orders. 
    
    \item \textbf{Shaped reward:} This reward extends the queue-based approach and introduces explicit penalties and incentives for charging decisions. 
    The agent is penalized for initiating charging when no stations are available and for charging actions that result in situations where no AMRs remain free while there are queued orders. 
    Conversely, rewards are given for charging actions taken when there are no queued orders and when a free charging station is available.
\end{itemize}

Where possible, we aim to normalize the reward components to the range $[0, 1]$. 
To that end, we divide the service time and queued retrieval orders by defined upper limits. 
The number of free AMRs is normalized by the total number of AMRs. 

% Motivate Interrupt vs not Interrupt during training
The final two configuration aspects we investigate pertain to the use of the previously introduced \textit{Interrupt} heuristic. 
We consider both training and evaluation with and without \textit{Interrupt} enabled.  
We compare the four different configurations shown in Table \ref{tab:rlconfigs}. 
The first three configurations differ in their reward formulations. 
\textit{Basic1} and \textit{Basic2} use the service-time and queue-based reward, respectively, incorporating minimal domain knowledge. 
\textit{LightShaped} uses the composite reward, including the free AMR component. 
All three configurations use $A_{full}$ and exclude \textit{Interrupt} during training. 
The fourth configuration, \textit{FullyShaped}, combines the shaped reward with the reduced action space $A_{binary}$. 
Additionally \textit{Interrupt} is enabled during training.
During evaluation, we assess configurations both with and without \textit{Interrupt}.
\begin{table}[!h]
\vspace{-10pt}
\caption{\small Overview of the different RL-configurations.}
\setlength{\tabcolsep}{2pt}
\renewcommand{\arraystretch}{1.2}
\scriptsize
\begin{tabularx}{\textwidth}{
>{\raggedright\arraybackslash}p{2cm}CCCCC}
\toprule
\multirow{2}{*}{\textbf{}} & \textbf{Reward} & \textbf{Action Space} & \textbf{Interrupted Training} & \textbf{Interrupted Evaluation} \\[6pt]
\midrule
\textit{Basic1} & \makecell[c]{Service time-based} & \makecell[c]{$A_{full}$} & \makecell[c]{False} & \makecell[c]{True, False} \\
\textit{Basic2} & \makecell[c]{Queue-based} & \makecell[c]{$A_{full}$} & \makecell[c]{False} & \makecell[c]{True, False} \\
\textit{LightShaped} & \makecell[c]{Composite reward \\+ Free AMRs} & \makecell[c]{$A_{full}$} & \makecell[c]{False} & \makecell[c]{True, False} \\
\textit{FullyShaped} & \makecell[c]{Shaped reward} & \makecell[c]{$A_{binary}$} & \makecell[c]{True} & \makecell[c]{True, False} \\
\bottomrule
\end{tabularx}
\label{tab:rlconfigs}
\vspace{-10pt}
\end{table}

\paragraph{Model}
We use PPO, an on-policy deep actor-critic RL algorithm \cite{https://doi.org/10.48550/arxiv.1707.06347}, primarily chosen for its stable action masking implementation in \cite{stable-baselines3}. 
Another advantage of PPO is its use of a clipped objective function, which constrains policy updates and mitigates the instability typical of policy gradient methods, thereby promoting stable and reliable training.

%% file: content/5_experiments.tex
%==========================================================================================
\section{Experiments}
\label{sec:experiments}
In this section, we first present our setup, followed by an analysis of the training outcomes for the proposed configurations. 
This is followed by a comprehensive evaluation of the best models in terms of generalization capabilities and operational performance metrics.

\paragraph{Setup}
All experiments were carried out on an Intel(R) Xeon(R) w5-2445 CPU.
We use the open-source implementation of the masked PPO algorithm from \cite{stable-baselines3} applying the default parameters. 
A rollout buffer size of 2048 and a minibatch size of 64 were used. 
Both the policy and value networks use a multi-layer perceptron, each consisting of two layers with 64 neurons. 
The order dataset, containing 400,000 orders, was divided on a weekly basis.  %Epochen minibatches

Four randomly sampled weeks were used for training. 
Each training episode terminates once all orders within the selected week have been simulated. 
The sampling resulted in weeks 3, 4, 6 and 13 being selected, containing 34,610, 37,134, 36,668, and 18,833 orders, respectively. 
Training was limited to 4 million steps, with periodic evaluation every 200,000 steps to retain the best performing model. 
This periodic evaluation was carried out over all training data sets and averaged. 
The same simulation setting as in the baseline comparison was used, involving 3 charging stations and 40 AMRs. 

\begin{figure}[htb!]
    \vspace{-10pt}
    \centering
    \subfloat[\small Mean episode reward.]{
        \includegraphics[width=0.48\linewidth]{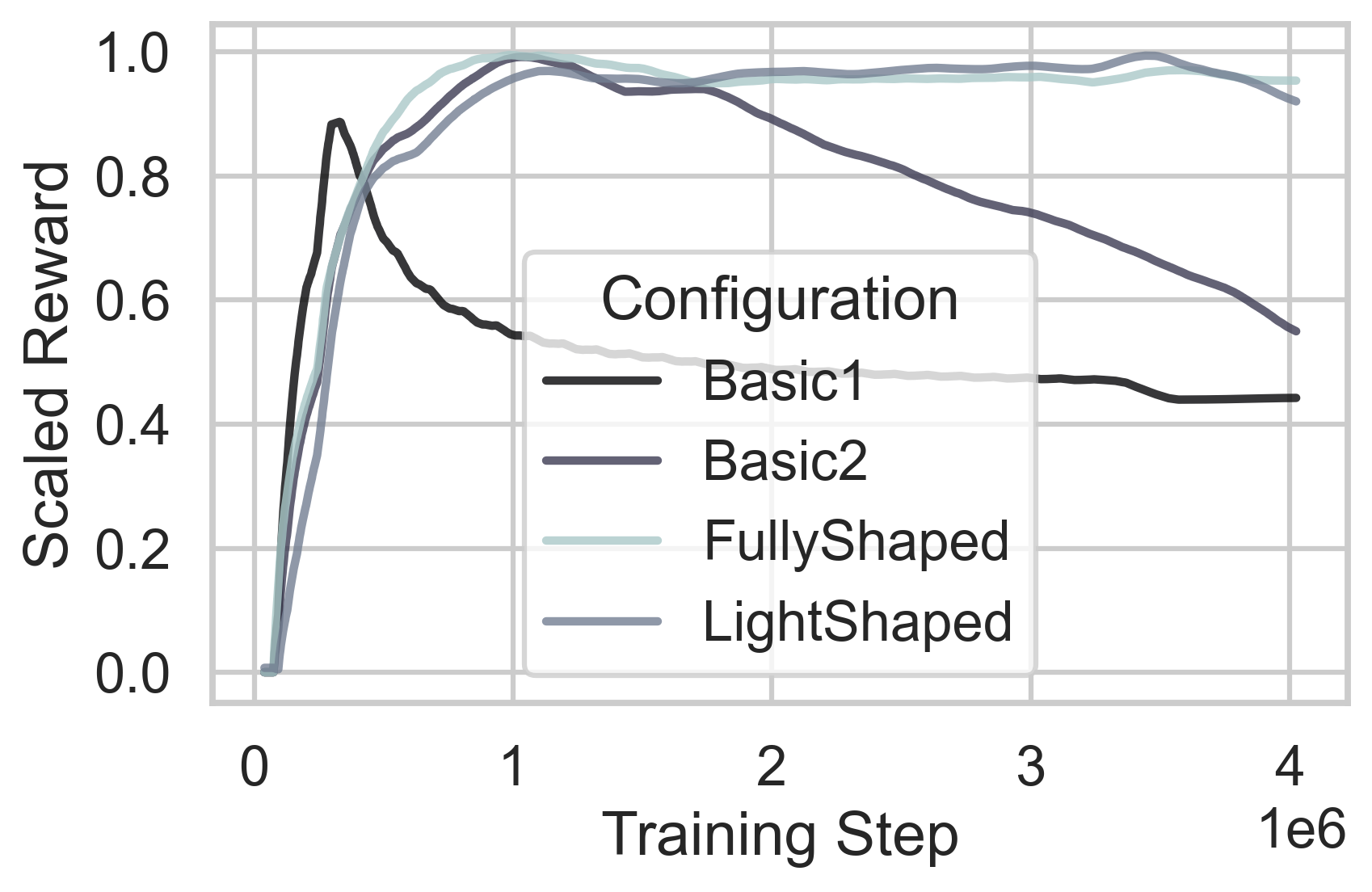}
        \label{fig:reward}
    }
    \subfloat[\small Entropy loss.]{
        \includegraphics[width=0.48\linewidth]{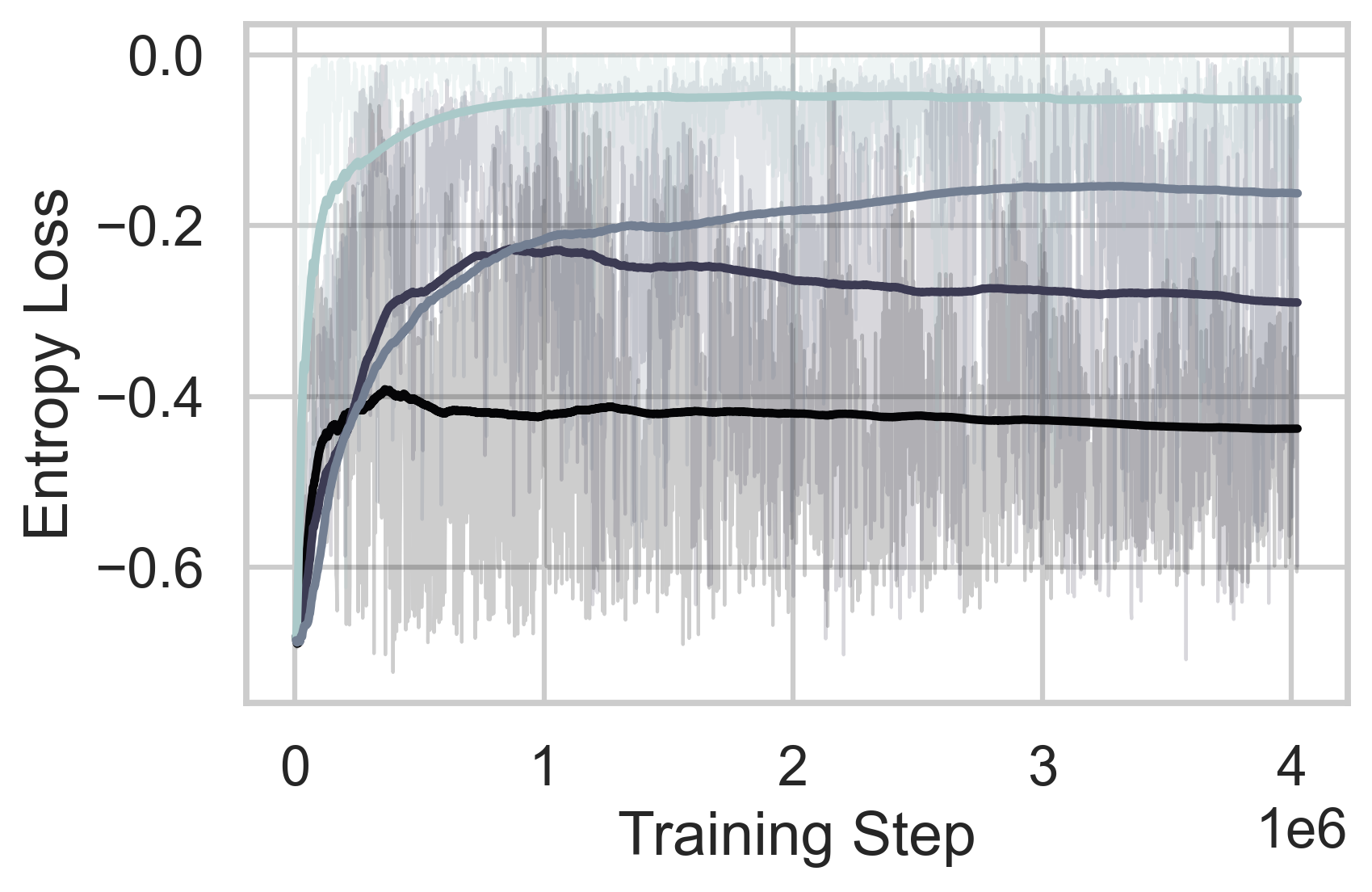}
        \label{fig:entr_loss}
    }
    \caption{\small Learning metric evolution during training for the proposed configurations.}
    \label{fig:learning_curves}
    \vspace{-10pt}
\end{figure}

\paragraph{Training}
The average episode reward for the different configurations is displayed in Figure \ref{fig:reward}. 
For a better visualization, we normalized the rewards to the range $[0, 1]$. 
As described by Schulman et al. in \cite{schulman2017proximal}, the policy entropy can be used to encourage exploration during training by incorporating an entropy bonus to the loss function. 
This additional loss is the negative mean entropy of the policy's action distribution across a batch of observations and is visualized for the different configurations in Figure \ref{fig:entr_loss}. 
Note that our training runs disregard the entropy term during policy update.
Nevertheless, the value is still useful as an indicator of exploration.

In terms of reward the \textit{Basic1} configuration initially performs well but declines after around 200,000 steps and never recovers. 
The reward curve for \textit{Basic2} improves over a longer duration begins to decline after one million steps. 
\textit{LightShaped} maintains strong performance for more than three million steps, with a decline observed after approximately 3.5 million steps. 
The \textit{FullyShaped} configuration yields the most stable reward curve, with an indication of convergence toward the end of training.
The entropy loss curves provide further insight into these dynamics. 
\textit{Basic1} and \textit{Basic2} exhibit poor convergence behavior, with persistently unstable entropy indicating ongoing exploration and a failure to form stable policies. 
This is consistent with their subpar reward performance. 
This suggests that the service-time- and queue-based rewards are not suited to effectively guide the agent toward reliable long-term strategies. 
\textit{FullyShaped}, by contrast, converges rapidly to a low-entropy policy, with the entropy loss approaching and remaining near zero. 
This indicates limited exploration. 
\textit{LightShaped} strikes a balance: entropy remains relatively high, supporting continued exploration, while gradually decreasing, reflecting policy stabilization. 
This controlled exploration aligns with its strong and sustained reward performance.
While \textit{Basic1} and \textit{Basic2} struggle with instability and performance degradation, the configurations with shaped reward functions \textit{LightShaped} and \textit{FullyShaped} demonstrate more robust learning dynamics. 
This suggests that incorporating more structured feedback into the reward function and reducing the action space enhances the agent's learning stability but may limit the exploration of the action space and encourage early deterministic policies, as can be seen from the entropy loss curve of \textit{FullyShaped}. 

\paragraph{Evaluation}
To assess the model generalization capabilities, we report the performance of the best models found during training, averaged over all non-training weeks (Table \ref{tab:charging_results}). 
We compare the average service time, the maximum and average number of queued retrieval orders, the average battery level, and the average distance traveled per AMR. 
The results are grouped based on the use of the \textit{Interrupt} heuristic, and 
we further include the top-performing heuristics from the baselines in Section \ref{sec:problem-setting}. 

\begin{table}[!h]
    \vspace{-10pt}
    \caption{\small Comparison of the best RL-models found during training. Results are grouped by use of \textit{Interrupt} and sorted by \textbf{Avg. Service Time (s)}. Best values are highlighted.}
\setlength{\tabcolsep}{2pt}
\renewcommand{\arraystretch}{1.2}
\scriptsize
\begin{tabularx}{\textwidth}{
>{\raggedright\arraybackslash}p{2.2cm}CCCCCC}
\toprule
\multirow{2}{*}{\textbf{Strategy}} & \textbf{Avg Service Time (s)} & \textbf{Max Retrieval Queue} & \textbf{Mean Retrieval Queue} & \textbf{Mean Battery Level} & \textbf{Mean Travel Distance (km/AMR)} \\[6pt]
\midrule
\multicolumn{6}{l}{\textbf{Interrupted False}} \\\midrule
\textit{LightShaped} & \makecell[c]{588.84} & \makecell[c]{\textbf{290}} & \makecell[c]{19.94} & \makecell[c]{61.99} & \makecell[c]{139.77} \\
\textit{Basic2} & \makecell[c]{675.17} & \makecell[c]{\textbf{290}} & \makecell[c]{22.06} & \makecell[c]{66.78} & \makecell[c]{137.35} \\
\textit{HighLow 40\% Th} & \makecell[c]{702.12} & \makecell[c]{\textbf{290}} & \makecell[c]{22.66} & \makecell[c]{52.34} & \makecell[c]{131.56} \\
\textit{Fixed 60\% Th} & \makecell[c]{714.87} & \makecell[c]{\textbf{290}} & \makecell[c]{22.20} & \makecell[c]{45.42} & \makecell[c]{131.66} \\
\textit{FullyShaped} & \makecell[c]{719.70} & \makecell[c]{346} & \makecell[c]{24.10} & \makecell[c]{79.79} & \makecell[c]{145.38} \\
\textit{Opportunity} & \makecell[c]{1223.09} & \makecell[c]{367} & \makecell[c]{37.64} & \makecell[c]{78.28} & \makecell[c]{142.24} \\
\textit{Basic1} & \makecell[c]{4897.74} & \makecell[c]{950} & \makecell[c]{149.99} & \makecell[c]{57.62} & \makecell[c]{163.82} \\
\bottomrule
\multicolumn{6}{l}{\textbf{Interrupted True}} \\\midrule
\textit{FullyShaped} & \makecell[c]{\textbf{581.85}} & \makecell[c]{\textbf{290}} & \makecell[c]{19.73} & \makecell[c]{64.27} & \makecell[c]{144.83} \\
\textit{LightShaped} & \makecell[c]{582.33} & \makecell[c]{\textbf{290}} & \makecell[c]{\textbf{19.52}} & \makecell[c]{56.14} & \makecell[c]{139.79} \\
\textit{Basic2} & \makecell[c]{601.67} & \makecell[c]{\textbf{290}} & \makecell[c]{20.23} & \makecell[c]{60.74} & \makecell[c]{137.97} \\
\textit{Fixed 90\% Th} & \makecell[c]{628.33} & \makecell[c]{296} & \makecell[c]{20.84} & \makecell[c]{49.42} & \makecell[c]{131.37} \\
% \textit{HighLow 40\% Th} & \makecell[c]{646.18} & \makecell[c]{337} & \makecell[c]{21.12} & \makecell[c]{45.82} & \makecell[c]{132.47} \\
\textit{HighLow 90\% Th} & \makecell[c]{685.00} & \makecell[c]{317} & \makecell[c]{22.39} & \makecell[c]{48.87} & \makecell[c]{136.01} \\
\textit{Opportunity} & \makecell[c]{709.55} & \makecell[c]{\textbf{290}} & \makecell[c]{23.85} & \makecell[c]{64.11} & \makecell[c]{144.42} \\
\textit{Basic1} & \makecell[c]{5027.13} & \makecell[c]{935} & \makecell[c]{156.72} & \makecell[c]{51.97} & \makecell[c]{148.87} \\
\bottomrule
\end{tabularx}
\label{tab:charging_results}
\vspace{-10pt}
\end{table}

Table \ref{tab:charging_results} illustrates how reward structure, action space design, and the \textit{Interrupt} heuristic influence AMR charging performance.
We can note that with and without the use of \textit{Interrupt} we can find RL-based charging strategies that outperform the best baseline in terms of average service times and number of queued retrieval orders. 
%Compared to \textit{HighLow} we can see that the mean travel distance per AMR is increased for the RL strategies. 
%This is probably due to shorter charging cycles that require AMRs to travel to charging stations more frequently.
Compared to \textit{HighLow}, RL strategies show increased average travel distance per AMR, likely due to shorter charging cycles that necessitate more frequent trips to charging stations.
The \textit{FullyShaped}  configuration, which combines a shaped reward with a reduced action space and domain knowledge, achieves the lowest average service time (581.85s) when interruption is enabled. 
However, \textit{LightShaped} performs best in the non-interrupted case, achieving an average service time of 588.84s. 
Furthermore, when interruption is enabled, \textit{LightShaped} results in lower average retrieval queues (19.52) compared to \textit{FullyShaped}  (19.73).
Both \textit{Basic1} and \textit{Basic2}, which use service-time and queue-based rewards, perform notably worse.
Most notably, \textit{Basic1}, which directly optimizes for service time, performs worst overall, with average service times of 5027.13s (with interruption) and 4897.74s (without).
\textit{Basic2} that only indirectly affects service times performs significantly better with average service times of 675.17s when not interrupted.  
Both \textit{Basic2} and \textit{LightShaped} can be considerably enhanced with \textit{Interrupt}, reducing both service times and queued retrieval orders. 
The \textit{FullyShaped} model, trained with interruption enabled, performs significantly worse when evaluated without it --- service times rise to 719.70s, and the queued retrieval orders peak at 346. 
This suggests that \textit{FullyShaped} fails to generalize as well as the more adaptable \textit{LightShaped} model.
%

%% file: content/6_conclusion.tex
%==========================================================================================
\section{Conclusion}
\label{sec:conclusion}
In this work, we extended an open-source simulation framework to support experiments on different AMR charging strategies in an ABSW setting. 
We also presented an MDP to \textit{jointly} handles both the decision to charge and the duration of charging. 
Finally, we evaluated different MDP configurations and showed that RL-based charging strategies can outperform common heuristics. 
Our study highlights that the choice of reward function, action space, and use of heuristics can have a significant impact on the learning stability and performance of RL agents.  
Using a reward function without domain knowledge, such as service times, in combination with a broad action space yielded the worst results.   
The best performance was achieved using a shaped reward, a reduced action space, and a simple heuristic rule to interrupt charging.
The inability of the model to perform well without this mechanism underscores the dangers of overfitting to specific problem settings.
The most balanced setup used a multi-objective reward function with domain knowledge. 

Given our findings, future research should explore more systematic approaches to reward shaping and validation. 
To improve realism, future work should incorporate more detailed battery models that account for degradation and non-linear charging behavior.
Additionally, further work is needed to better isolate scenarios where battery management is the primary performance bottleneck, as opposed to other system aspects such as layout design or order sequence. 
A more thorough and systematic evaluation of algorithms and parameters, e.g. learning rate, is also essential to fully understand the potential of the promising RL designs proposed in this work.

\subsubsection{\ackname}
This research was funded by the European Union - NextGenerationEU - funding code 13IK032I